\documentclass[10pt,twocolumn,letterpaper]{article}

\usepackage{cvpr}              
\usepackage[accsupp]{axessibility} 
\usepackage[dvipsnames]{xcolor}

\definecolor{cvprblue}{rgb}{0.21,0.49,0.74}
\usepackage{amsmath}
\usepackage[pagebackref,breaklinks,colorlinks,citecolor=cvprblue]{hyperref}

\usepackage{xcolor}
\usepackage{booktabs}
\usepackage{multirow}
\usepackage{makecell}
\usepackage{algorithm}  
\usepackage{algpseudocode}  
\usepackage{amsmath}  
\usepackage{setspace}
\usepackage{adjustbox}
\def\paperID{9436} 
\def\confName{CVPR}
\def\confYear{2024}

\begin{document}
\title{LayoutLLM: Layout Instruction Tuning with Large Language Models for Document Understanding}


\author{
Chuwei Luo$^{1*}$, 
Yufan Shen$^{12*}$, 
Zhaoqing Zhu$^{1*}$, 
Qi Zheng$^1$, 
Zhi Yu$^2$, 
Cong Yao$^1$ \\
$^1$Alibaba Group, 
$^2$Zhejiang University\\
{\tt\small \{luochuwei,zzhaoqing.z,zhengqisjtu,yaocong2010\}@gmail.com}\\
\small{\texttt{\{syficy,yuzhirenzhe\}@zju.edu.cn}}
}
\maketitle
\def\thefootnote{*}\footnotetext{Equal contribution.}
\def\thefootnote{\arabic{footnote}}

\begin{abstract}
Recently, leveraging large language models (LLMs) or multimodal large language models (MLLMs) for document understanding has been proven very promising. However, previous works that employ LLMs/MLLMs for document understanding have not fully explored and utilized the document layout information, which is vital for precise document understanding. In this paper, we propose LayoutLLM, an LLM/MLLM based method for document understanding. The core of LayoutLLM is a layout instruction tuning strategy, which is specially designed to enhance the comprehension and utilization of document layouts. The proposed layout instruction tuning strategy consists of two components: Layout-aware Pre-training and Layout-aware Supervised Fine-tuning. To capture the characteristics of document layout in Layout-aware Pre-training, three groups of pre-training tasks, corresponding to document-level, region-level and segment-level information, are introduced. Furthermore, a novel module called layout chain-of-thought (LayoutCoT) is devised to enable LayoutLLM to focus on regions relevant to the question and generate accurate answers. LayoutCoT is effective for boosting the performance of document understanding. Meanwhile, it brings a certain degree of interpretability, which could facilitate manual inspection and correction. Experiments on standard benchmarks show that the proposed LayoutLLM significantly outperforms existing methods that adopt open-source 7B LLMs/MLLMs for document understanding.

\end{abstract}
\vspace{-6mm}

\section{Introduction}
\label{sec:intro}

Document AI~\cite{cui2021document}, including its document understanding tasks such as document VQA~\cite{mathew2021docvqa,tanaka2021visualmrc} and document visual information extraction~\cite{jaume2019funsd,park2019cord,Huang_2019}, is currently a hot topic in both academia and industry. In recent years, document pre-trained models~\cite{xu2020layoutlm,xu2020layoutlmv2,huang2022layoutlmv3,li2021selfdoc,li2021structurallm,li2021structext,appalaraju2021docformer,gu2022unified,wang2022lilt,gu2022xylayoutlm,hong2022bros,yu2023structextv,peng2022ernielayout,Luo_2023_CVPR,da2023vision} have achieved excellent performance in document AI downstream tasks. However, due to the necessity for fine-tuning on corresponding downstream task data, it is challenging to directly adapt such pre-trained models for \textit{zero-shot} document understanding. 
In this paper, \textit{zero-shot} refers to not using training sets of downstream tasks.

\begin{figure}[tp]
    \includegraphics[width=.95\linewidth]{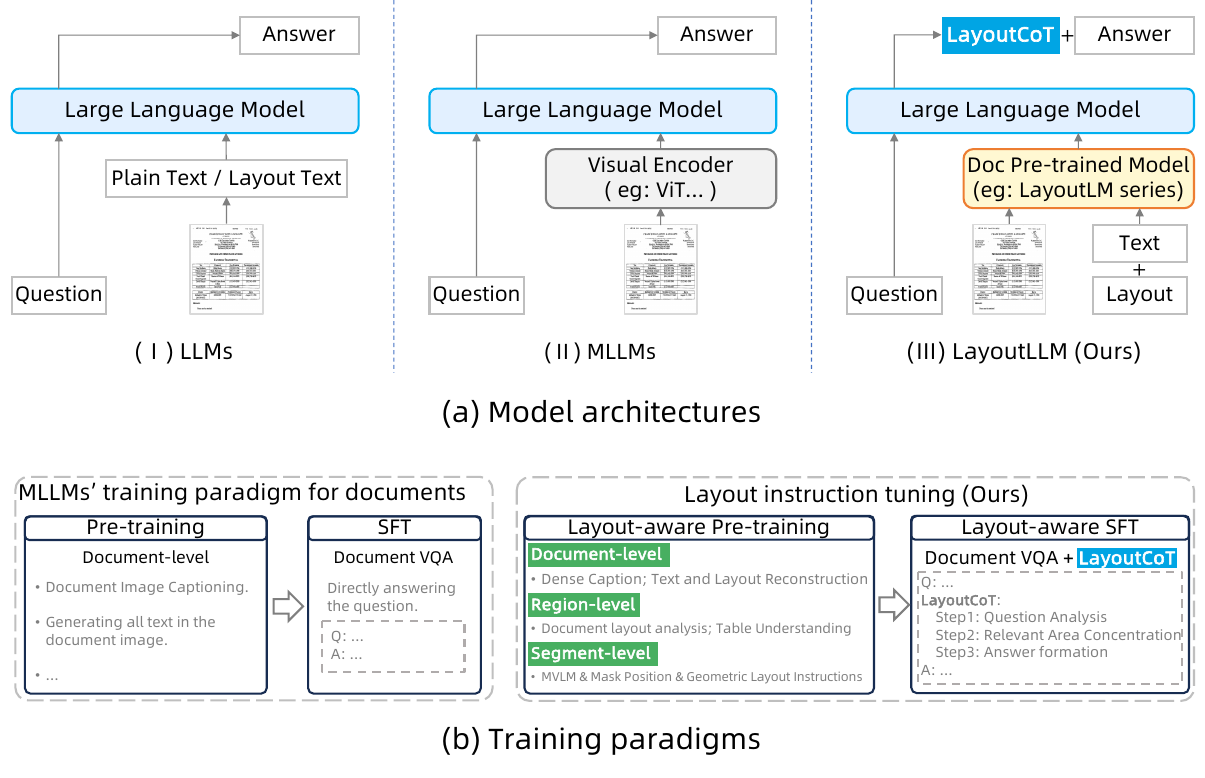}
    \caption{
    LLMs/MLLMs for document understanding.
    The LayoutLLM is an LLM/MLLM based method that integrates a document pre-trained model as encoder. It is trained by the newly proposed layout instruction tuning strategy which consists of Layout-aware Pre-training and Layout-aware Supervised Fine-tuning.
    }
    \label{intro_case}
    \vspace{-6mm}
\end{figure}

Recently, large language models (LLMs) such as ChatGPT~\cite{chatgpt_webpage} and LLaMA~\cite{touvron2023llama,touvron2023llama2}, or multimodal large language models (MLLMs) like GPT-4V~\cite{openai2023gpt4,2023GPT4VisionSC,yang2023dawn}, have shown remarkable zero-shot capabilities across various applications. For Document AI, as shown in~\cref{intro_case} (a), (I) directly prompting LLMs with document text~\cite{perot2023lmdx,he2023icl} and (II) training document-based MLLMs~\cite{zhang2023llavar,ye2023mplugdoc,bai2023qwen} have also achieved promising results under the zero-shot setting~\cite{perot2023lmdx,zhang2023llavar,ye2023mplugdoc,bai2023qwen}.

It is widely accepted that document layout information is vital for document understanding~\cite{xu2020layoutlm,xu2020layoutlmv2,huang2022layoutlmv3,powalski2021going,li2021selfdoc,li2021structurallm,li2021structext,appalaraju2021docformer,gu2022unified,wang2022lilt,gu2022xylayoutlm,hong2022bros,yu2023structextv,peng2022ernielayout,Luo_2023_CVPR,da2023vision}. However, it is difficult to convey document layout information by directly feeding text to LLMs. As \cref{intro_case}(a)(I) shows, representing documents as either flattened plain text or layout text such as text with coordinates~\cite{perot2023lmdx,he2023icl,shi2016end, zhou2017east} is often used for LLMs. Flattened plain text completely excludes any layout information of the document~\cite{xu2020layoutlm}. Additionally, as \cref{table:main_results} shows, using layout text that represents both textual and layout information as inputs for LLMs does not guarantee LLMs can effectively comprehend this formatted text. 

Moreover, existing works that employ MLLMs for document understanding also have not fully explored the document layout information. Document-based MLLMs integrate visual models~\cite{vit,radford2021learning} with LLM~\cite{touvron2023llama,touvron2023llama2,alpaca,zheng2023judging} for document understanding. As \cref{intro_case}(b) shows, they are typically based on pre-training and supervised fine-tuning (SFT) on document datasets. In the pre-training stage, tasks such as image captioning~\cite{liu2023llava,zhang2023llavar,ye2023mplugdoc} or generating all text in a document as flattened plain text~\cite{zhang2023llavar,liao2022real, da2023multi} are commonly applied. Both these image captions~\cite{sidorov2020textcaps,schuhmann2022laion} and plain text only provide a brief representation and fail to capture the layout information of the document. So it is difficult for the model to learn document layout in the existing pre-training stage. In the SFT stage, document-related VQA or information extraction data~\cite{ye2023mplugdoc,bai2023qwen} is often used. The answers are directly provided during SFT, lacking explicit learning about document layout. In summary, current approaches using plain or layout text to prompt LLMs and training document-based MLLMs have not effectively captured layout information, limiting their zero-shot document understanding capability. Therefore, for better document understanding with the power of LLMs, it is necessary to investigate how to effectively incorporate layout information into LLMs.

To this end, we propose \textbf{\textit{LayoutLLM}}, an LLM/MLLM based method for document understanding, in which a layout instruction tuning strategy is designed to enhance the comprehension of document layouts. Different from the existing MLLMs that use a general visual pre-trained model~\cite{vit,radford2021learning} as the encoder, we integrate document pre-trained models~\cite{xu2020layoutlm,xu2020layoutlmv2,huang2022layoutlmv3,li2021selfdoc,li2021structurallm,appalaraju2021docformer,gu2022unified,wang2022lilt,hong2022bros,Luo_2023_CVPR} as the encoder in order to better leverage the model's foundational understanding capability for documents. The proposed layout instruction tuning consists of two stages: layout-aware pre-training and layout-aware supervised fine-tuning (SFT). Due to the complex nature of documents in real-world scenarios, encompassing rich textual content and diverse layout structures, achieving a thorough understanding involves not only capturing the document's fundamental content at global but also delving into local details. In the layout-aware pre-training stage, to ensure the model learns not only the global information of documents but also detailed information at different hierarchical levels, three groups of different level pre-training tasks are proposed: document-level, region-level, and segment-level. All the proposed pre-training tasks are unified in the format of instruction tuning.


Furthermore, in the layout-aware SFT stage, to enhance the model's comprehension and utilization of layout information for question answering, a novel strategy called \textit{LayoutCoT} is proposed, motivated by the chain-of-thought (CoT)~\cite{wei2022chain,kojima2022large} ability in LLMs.
Unlike existing methods that are directly supervised by the answer to the document understanding question, \textit{LayoutCoT} consists of three successive steps: \textit{Question Analysis}, \textit{Relevant Area Concentration}, and \textit{Answer Formation}. 
Through these steps, the model gains a deeper understanding of the questions, becomes capable of focusing the the relevant areas instead of searching answers in the entire document and can leverage the specific characteristics of identified areas (such as tables, paragraphs, etc.) to accurately infer the answers. It not only brings a certain degree of interpretability, but also provides a feasible way for manual intervention or correction of model results. Extensive zero-shot experiments on five widely-used document understanding benchmarks demonstrate the effectiveness of the proposed LayoutLLM.

Our contributions are summarized as follows:

\begin{itemize} \setlength{\itemsep}{0pt}
    \item[1)] To better learn document layouts from global to local in layout-aware pre-training, three groups of different level pre-training tasks, which are all implemented through instruction tuning, are proposed.    
    \item[2)] A novel \textit{LayoutCoT} strategy is proposed to achieve layout-aware supervised fine-tuning. It enables LayoutLLM to focus on the relevant document area and leverage the region's features to generate accurate answers, exhibiting a certain degree of interpretability.
    \item[3)]  Experimental results on zero-shot document understanding tasks show that the proposed LayoutLLM significantly outperforms existing methods that adopt LLMs/MLLMs for document understanding, demonstrating the great potential of document layout modeling.
\end{itemize}

\section{Related Works}
\label{sec:related_works}

\noindent{\bf{Pre-trained models for document understanding.}}
Document pre-trained models have demonstrated the effectiveness of layout information in document understanding~\cite{xu2020layoutlm,xu2020layoutlmv2,huang2022layoutlmv3,powalski2021going,li2021selfdoc,li2021structurallm,li2021structext,appalaraju2021docformer,gu2022unified,wang2022lilt,gu2022xylayoutlm,hong2022bros,yu2023structextv,peng2022ernielayout,Luo_2023_CVPR,da2023vision,kim2022ocr,davis2022end,lee2023pix2struct,cao2023attention}.
As a pioneer, LayoutLM~\cite{xu2020layoutlm} is the first to encode spatial coordinates of text for layout representation learning in pre-training documents.
The following works~\cite{xu2020layoutlmv2,huang2022layoutlmv3,li2021selfdoc,li2021structurallm,li2021structext,appalaraju2021docformer,gu2022unified,wang2022lilt,gu2022xylayoutlm,hong2022bros,yu2023structextv,peng2022ernielayout,Luo_2023_CVPR,da2023vision} then joint text, layout and images in document pre-training by combining visual models as document image encoders with the text and layout transformers, and various works~\cite{kim2022ocr,davis2022end,lee2023pix2struct,cao2023attention} start to explore pre-training end-to-end models for document understanding.
These studies have achieved significant advancements in document understanding by exploring various model architectures~\cite{powalski2021going,li2021selfdoc,li2021structext,appalaraju2021docformer,gu2022unified,wang2022lilt,gu2022xylayoutlm,kim2022ocr,davis2022end,yu2023structextv,peng2022ernielayout,da2023vision} and attention mechanisms~\cite{xu2020layoutlmv2,huang2022layoutlmv3,hong2022bros} for modeling layout information.
These methods also have proposed layout-related pre-training tasks that have been demonstrated to be highly effective in document understanding tasks.
For instance, tasks like masked vision-language modeling~\cite{gu2022xylayoutlm,xu2020layoutlmv2,huang2022layoutlmv3}, where the model is required to generate the original text corresponding to the randomly masked text in the document; position masking~\cite{luo2022bivldoc,tu-etal-2023-layoutmask}, involving the randomly position masking and subsequent recovery of position information in the document; geometric pre-training~\cite{li2021structext,Luo_2023_CVPR}, focusing on learning direction, distance, etc.; and layout-aware generation tasks~\cite{lee2023pix2struct,cao2023attention}, aiming to make the model generate structured text with layout information.
However, due to the necessity of fine-tuning with annotated data for downstream tasks, these efforts face challenges in extending to zero-shot document understanding.


\noindent{\bf{LLMs/MLLMs for document understanding.}}
Recently, LLMs like ChatGPT~\cite{chatgpt_webpage} and MLLMs like GPT-4~\cite{openai2023gpt4,yang2023dawn} have demonstrated remarkable zero-shot performance across a wide range of NLP/CV tasks. 
Leveraging LLMs/MLLMs for zero-shot document understanding has also shown promising progress~\cite{perot2023lmdx,liu2023hidden,zhang2023llavar,ye2023mplugdoc,bai2023qwen,shi2023exploring}.
\citet{perot2023lmdx} explore the use of LLMs for document visual information extraction, emphasizing the importance of the document layout.
LLaVAR~\cite{zhang2023llavar} which extends LLaVA~\cite{liu2023llava,liu2023improvedllava} to the document domain is pre-trained by generating plain text in the document image. During SFT, it is trained by document-related instructions which are generated by GPT-4.
Expanding on mPLUG-Owl~\cite{ye2023mplugowl}, mPLUG-DocOwl~\cite{ye2023mplugdoc} is trained using publicly available datasets for document understanding. It includes tasks like document-level image captioning, direct information extraction, and direct document VQA.
Moreover, Qwen-VL~\cite{bai2023qwen} proposes a general MLLM that performs well on document understanding tasks, utilizing document-level pre-training and direct VQA for SFT.
Though existing LLMs/MLLMs have shown promising results in document understanding, their limited focus on document layout in pre-training and SFT has hindered their ability to achieve higher accuracy in zero-shot document understanding and better interpretability.


\section{LayoutLLM}

LayoutLLM is an LLM/MLLM based method that incorporates document pre-trained models for document understanding.
To enhance the document layout comprehension in LayoutLLM, a novel layout instruction tuning strategy is proposed, which consists of two stages: layout-aware pre-training and layout-aware supervised fine-tuning (SFT).

\subsection{Model Architecture}

\begin{figure}[tp]
    \centering
    \setlength{\abovecaptionskip}{1mm}
    \includegraphics[width=.75\linewidth]{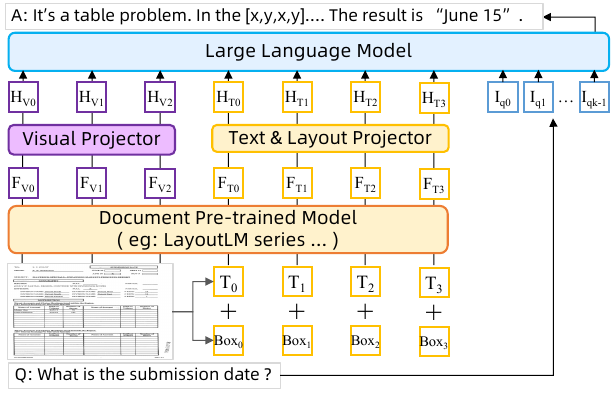}
    \caption{Overall architecture of LayoutLLM.}
    \label{layoutllm_model}
    \vspace{-5mm}
\end{figure}

The overall architecture of LayoutLLM is shown in \cref{layoutllm_model}. 
In LayoutLLM, given an input document image and its corresponding text and layout information, the document pre-trained model encoder is required to obtain the multimodal document features.
Then, these features are encoded by multimodal projectors, and together with the instruction embeddings, fed into the LLM to generate the final results.

\noindent{\bf{Document pre-trained model encoder.}} To leverage the foundational document comprehension capability of document pre-trained models, in this work, we utilize LayoutLMv3~\cite{huang2022layoutlmv3}, a widely-used document pre-training model, as our basic document encoder.
The document image, text, and layout are initially inputted into the document pre-trained model ($DocPTM$). They are then encoded by the $DocPTM$ to obtain the corresponding features as follows:
\begin{equation}
    \begin{aligned}
    F_V, F_T &= {DocPTM}(V, T, Box)
    \end{aligned}
    \label{encode_func}
\end{equation}
where $V$ represents the document image, $T=T_{0:n-1}$ and $Box=Box_{0:n-1}$ indicate the text sequences in the document and their corresponding bounding-box coordinates respectively. After being encoded by the $DocPTM$, the visual features of the document $F_V=F_{V_0:V_{m-1}} \in \mathbb{R}^{d_0}$ and the text layout features $F_T=F_{T_0:T_{n-1}} \in \mathbb{R}^{d_0}$ are acquired. $m$ signifies the number of visual features and $n$ represents the number of tokens contained in the document. $d_0$ denotes the dimension of $DocPTM$ feature space.

\noindent{\bf{Multimodal projectors.}} To project multi-modality features from $DocPTM$ into the LLM's embedding space, inspired by the simple yet effective projector design in LLaVA~\cite{liu2023llava,liu2023improvedllava}, two different Multilayer Perceptrons (MLPs) are used as visual projector and text \& layout projector respectively. Formally, the projected features can be obtained by:
\begin{align}
    & H_V=P_V(F_V) \\
    & H_T=P_T(F_T)
    \label{project_func}
\end{align}
where $H_V=H_{V_0:V_{m-1}} \in \mathbb{R}^{d_1}$ is the feature encoded by the visual projector, $H_T=H_{T_0:T_{n-1}} \in \mathbb{R}^{d_1}$ is the feature encoded by the text \& layout projector, and $d_1$ is the dimensional of the LLM embedding features.

\noindent{\bf{Large language model.}} 
Finally, the $H_V$, $H_T$ and the embedding of the question's instruction text, $I_q=I_{q_{0:l_q-1}}$, are inputted together into the LLM, generating the target answer $I_a=I_{a_{0:l_a-1}}$. 
$l_q$ and $l_a$ represent the length of the question's instruction text and the answer text, respectively.


\subsection{Layout Instruction Tuning}

\begin{figure*}[ht]\centering
     \includegraphics[width=0.99\textwidth]{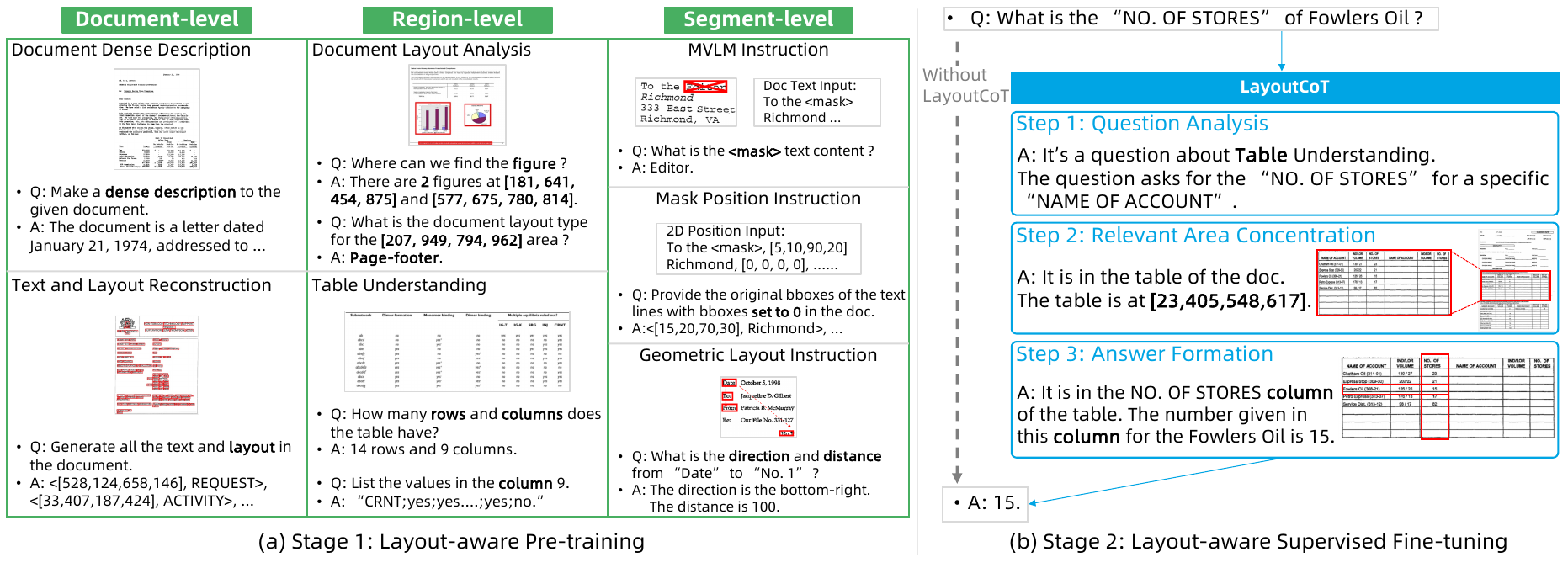} 
    \vspace{-1mm}
    \caption{
        Overview of the Layout Instruction Tuning. (a) Document-level, region-level, and segment-level pre-training tasks, unified in instruction tuning format, are introduced. (b) A novel module called \textit{LayoutCoT} is designed to enable LayoutLLM to focus on regions relevant to the question and generate accurate answers through three intermediate steps.
        %
        }
     \label{fig:layout_instruct}
    \vspace{-2mm}
\end{figure*}

The LayoutLLM model is trained using the layout instruction tuning, which consists of two stages: layout-aware pre-training and layout-aware SFT. 

\subsubsection{Layout-aware Pre-training}
The goal of pre-training the LayoutLLM is to enhance the model's comprehensive understanding of documents at different levels through layout learning, rather than only focusing on global document-level understanding like existing MLLM methods~\cite{ye2023mplugdoc,zhang2023llavar}.
To this end, during the pre-training stage, three different level pre-training strategies are simultaneously applied to the LayoutLLM, namely document-level, region-level, and segment-level.


\noindent{\bf{Document-level}} To enable the model to possess fundamental global document understanding, pre-training tasks, namely Document Dense Description (DDD) and Text and Layout Reconstruction (TLR), are proposed. 
As ~\cref{fig:layout_instruct} (a) shows, like the image caption task, the DDD task requires the model to learn to describe the input document.
Moreover, in the DDD task, the descriptions for document images are more detailed.
For instance, in the document image caption data used for the LLaVAR~\cite{zhang2023llavar} pre-training, captions contain an average of 36.27 words, while in the proposed dataset of the DDD task, the description contains an average of 373.25 words.
Through the DDD task, the model can obtain basic document-level information, such as document type and detailed content.
The TLR task aims to reconstruct the complete text and layout information of a document and output it in the format  ``\textless \{box\}, \{text\} \textgreater".
The TLR task aligns the text and layout embeddings output from DocPTM with the LayoutLLM's LLM space.
Consequently, it enables the LLM in LayoutLLM to comprehend the text and layout information contained in the documents.

\noindent{\bf{Region-level}} The information contained in specific regions of a document, such as the titles, figures, tables, is essential for document understanding~\cite{binmakhashen2019document,li2020docbank,zhong2019publaynet,cheng2023m6doc,da2023vision}. These regions serve as important characteristics that differentiate a document from plain text in natural language. For the LayoutLLM to achieve the basic region-level understanding, two pre-training tasks, namely Document Layout Analysis (DLA) and Table Understanding (TU), are utilized.
The DLA task is achieved in two ways as shown in ~\cref{fig:layout_instruct}. 
One involves locating the layout region based on the layout type, the other involves identifying the type of a given area.
Furthermore, the table region differs from other regions in that it requires additional focus on 2D layout understanding.
The TU task enables the model to understand the basic row and column information in the table region of a document. As shown in ~\cref{fig:layout_instruct}, the TU task includes instruction tuning for the number of rows and columns, logical coordinates, and the content within rows and columns.

\noindent{\bf{Segment-level}} 
Early works on document pre-trained models~\cite{gu2022xylayoutlm,xu2020layoutlmv2,huang2022layoutlmv3,luo2022bivldoc,tu-etal-2023-layoutmask,li2021structext,Luo_2023_CVPR} have demonstrated the effectiveness of segment-level document pre-training tasks to document layout understanding ability, such as masked vision language modeling (MVLM)~\cite{gu2022xylayoutlm,xu2020layoutlmv2,huang2022layoutlmv3}, position masking~\cite{luo2022bivldoc,tu-etal-2023-layoutmask}, and geometric pre-training~\cite{li2021structext,Luo_2023_CVPR}.
Inspired by these works, to make LayoutLLM have segment-level layout understanding, these tasks are transformed into instruction formats for pre-training as ~\cref{fig:layout_instruct}(a) shows. 
For the MVLM instructions, random masking of the text input to LayoutLLM is performed, and the model is instruction tuned by asking the masked words and answering them.
For mask position instruction, the layout information (coordinates) to a specific text line, when input to LayoutLLM, is randomly set to 0.
The instruction is constructed by asking about the text line with zeroed coordinates and requesting the model to respond with the original coordinates with text content.
For geometric layout instruction, text lines are randomly selected, and an instruction is constructed by asking questions about the direction and distance between them.

\subsubsection{Layout-aware Supervised Fine-tuning}
\label{sec:layoutsft}
In the SFT stage of existing document-based MLLMs, models are directly supervised by the answer to the document understanding instructions.
Consequently, these methods lack explicit learning of document layout which is crucial for document understanding.
Considering this limitation, and inspired by previous works related to chain-of-though (CoT)~\cite{kojima2022large,wei2022chain}, which have shown that inferencing with intermediate steps can greatly enhance performance.
A novel module called \textit{LayoutCoT} is proposed, which incorporates the layout information into every intermediate step of CoT explicitly.
Meanwhile, by introducing the layout-aware intermediate steps, the answer process gains a certain degree of interpretability for LayoutLLM and also provides interactive correction possibilities based on \textit{LayoutCoT}.



\noindent{\bf{LayoutCoT Details.}} 
As shown in Fig. \ref{fig:layout_instruct}(b), the \textit{LayoutCoT} involves the following three intermediate steps: 

\noindent{\textit{Step 1: Question Analysis.}}
To effectively address a document understanding problem, analyzing the key characteristics of the question is very important.
Identifying the question type, such as table understanding or entity extraction from paragraphs, and assessing whether the question is a straightforward extraction query or a more complex reasoning problem, can help guide the direction for the subsequent inference process.
Therefore, to give basic guidance to the subsequent steps, the question analysis step is designed, encompassing an analysis of the question type from a layout perspective and a detailed understanding of the question itself.
Benefiting from the layout understanding ability by layout-aware pre-training, this step can extract the types and key information mentioned in the question, which are related to the specific characteristics of the document.
%

\noindent{\textit{Step 2: Relevant Area Concentration.}}
For most document understanding tasks, the entire document contains a large amount of irrelevant information that may confuse the model~\cite{cao2023attention}.
This step aims to focus on the relevant area and generate its location information, which is used to assist the model to accurately infer the answer.
Benefiting from the layout information conveyed by step 1 and the positioning capabilities learned from the region\&segment-level pre-training, the model can accurately generate the location of the relevant area.
For example in \cref{fig:layout_instruct}(b), according to the question type ``table" in step 1, the relevant ``table" can be located.
By guiding the model to focus on the relevant area, this step largely narrows the search scope, increasing the possibility of giving the right answer. Meanwhile, the location information provides a way for visual inspection and interactive correction (see~\cref{sec:interactive} for details).

%
\noindent{\textit{Step 3: Answer Formation.}}
Finally, the last step, the answer formation, provides explanations based on the layout characteristics of the relevant areas located in step 2 and key points analyzed in step 1 to get the final answer.
For example in \cref{fig:layout_instruct}(b), for a ``table'' type question, this step involves analyzing the row and column in the relevant table in step 2, and inferencing the answer step-by-step.
For a ``key-value" question, analyzing the keywords in concentrated areas can help get the final answer.
Analyzing answers in different ways based on the features of different layout regions not only improves the document understanding performance but also brings a certain level of interpretability.
\begin{figure}[tb]\centering
    \vspace{-1.5em}
        \begin{minipage}{0.99\linewidth}
\begin{algorithm}[H] 
    \footnotesize
    \caption{CONSTRUCT($\mathcal{D}$): LayoutCoT Construction.}  
    \label{alg:LayoutCoT_algorithm}
    \begin{algorithmic}[1]
    \Statex\hspace*{-\algorithmicindent} \textbf{Definition:} $\mathcal{H}$: Document HTMLs; $\mathcal{I}$: Document Images; $\mathcal{T}$: MRC Texts;  
    $\mathcal{R}$: Document Language Representation; $\mathcal{QA}$: QA pairs; $\mathcal{T}_c$: Text CoT; $\mathcal{L}_c$: LayoutCoT;  
    \Statex\hspace*{-\algorithmicindent} \textbf{Input:} Document Dataset $\mathcal{D}=\{\mathcal{D}_H$, $\mathcal{D}_I$, $\mathcal{D}_M\}$ \ \newline ($\mathcal{D}_H$=\{$\mathcal{H}$\}, $\mathcal{D}_I$=\{$\mathcal{I}$\}, $\mathcal{D}_M = \{\mathcal{T}, \mathcal{QA}$\});
    \Statex\hspace*{-\algorithmicindent} \textbf{Output:} Constructed Dataset $\mathcal{D}_c$
    \vspace{0.5em}
    \State \textbf{Procedure} CONSTRUCT($\mathcal{D}=\{\mathcal{D}_H, \mathcal{D}_I, \mathcal{D}_M\}$)
    \State  \hspace{1em} 1) $\mathcal{R}$ = \textit{getDocRep}($\mathcal{I}$) if $\mathcal{D}\subseteq\mathcal{D}_I$ else $\mathcal{H}$ if $\mathcal{D}\subseteq\mathcal{D}_H$ else pass;
    \State  \hspace{1em} 2) $\mathcal{QA}, \mathcal{T}_c$ = \textit{getQACoT}($\mathcal{D}$) if $\mathcal{D}\subseteq\mathcal{D}_M$ else \textit{getQACoTGPT}($\mathcal{R}$);
    \State  \hspace{1em} 3) $\mathcal{L}_c$ = \textit{getLayoutCoT}($\mathcal{T}_c$)
    \State  \hspace{1em} 4) if $\mathcal{D}\subseteq\{\mathcal{D}_H, \mathcal{D}_M\}$: $\mathcal{I}$ = \textit{Html2Img}($\mathcal{D}_H$ ? $\mathcal{H}$ : $\mathcal{T}$);
    \\
    \Return $\mathcal{D}_{c}$ $\leftarrow$ \{$\mathcal{I},\mathcal{QA},\mathcal{L}_c$\}

    \end{algorithmic}
\end{algorithm}
\end{minipage}
\label{fig:algorithm_example}
\vspace{-4mm}
\end{figure}

\noindent{\bf{LayoutCoT Construction.}} 
Given the need for both text and image annotations in constructing \textit{LayoutCoT}, manual labeling can be difficult.
\cref{alg:LayoutCoT_algorithm} proposes a manual-labeling-free method, generating \textit{LayoutCoT} data using public datasets with GPT (GPT-3.5 Turbo).
It involves representing document text and layout in a format understandable by GPT. GPT is then utilized to generate document-content-based QA and corresponding text CoT. Finally, use rules for transforming the text CoT to LayoutCoT.

Three types of publicly available document datasets are focused on: HTML documents ($\mathcal{D}_H$), image documents ($\mathcal{D}_I$), and text documents ($\mathcal{D}_M$) for machine reading comprehension (MRC).
The construction process is as follows:
%

\noindent 1) \textbf{Document Representation}:
To fully leverage the capabilities of GPT, it is crucial to ensure that the document content fed to GPT contains accurate layout information.
Since HTML is the formatted language that can represent documents accurately, $\mathcal{D}_H$ is represented using the original HTML.
By transforming HTML to PDF and using the PDF parser, the text and bounding-boxes are obtained.
For $\mathcal{D}_I$, the layout-aware text~\cite{he2023icl} is used.
The text and bounding-boxes in $\mathcal{D}_I$ are from the original dataset annotations.

\noindent 2) \textbf{QA\&Text CoT Generation}: 
The language representation $\mathcal{R}$ for the document is employed for prompting GPT to generate QA pairs $\mathcal{QA}$ with text CoTs $\mathcal{T}_c$. In addition, the $\mathcal{D}_M$ includes the QA pairs and reasoning process, thereby directly reusing the $\mathcal{QA}$ and manually organizing $\mathcal{T}_c$.
The generated $\mathcal{T}_c$ includes the step 1 (question analysis) and step 3 (answer formation) for LayoutCoT, and locates all relevant sentences in the document for $\mathcal{QA}$.

\noindent 3) \textbf{LayoutCoT Generation}: 
The step 1 \& 3 in $\mathcal{T}_c$ are used as the step 1 \& 3 in $\mathcal{L}_c$.
To construct the step 2 (relevant area concentration) of $\mathcal{L}_c$, the union bounding-box of all located relevant sentences in $\mathcal{T}_c$ are taken as the relevant area.

\noindent 4) \textbf{Document Images Generation}: For $\mathcal{D}_H$ and $\mathcal{D}_M$, the HTMLs and MRC text are converted to images.
Overall, the document images $\mathcal{I}$, generated QA pairs $\mathcal{QA}$ and LayoutCoTs $\mathcal{L}_c$ constitute the final LayoutCoT dataset $\mathcal{D}_{c}$.

\section{Experiments}



\subsection{Dataset Collection}
\label{sec:data_collection}

\noindent{\bf{Layout-aware pre-training data}} of LayoutLLM is from publicly available document understanding datasets.
It does not incorporate any data from the training, validation, and test sets of downstream benchmarks.
Region-level pre-training tasks, most document-level and segment-level tasks are self-supervised.
Thus, only document images and images converted from PDFs in the original datasets, along with the corresponding OCR or text-layout results from PDF parsing, are needed.
For these tasks, data is randomly sampled from PubLayNet~\cite{zhong2019publaynet}, DocLayNet~\cite{Pfitzmann_2022}, Docbank~\cite{li2020docbank}, RVL-CDIP~\cite{harley2015evaluation}, and DocILE~\cite{2023docile}.
Particularly, data for document dense description is from inputting the document text content into GPT-3.5 Turbo, prompting it to generate an average of 373.25 words document dense descriptions.
For the region-level tasks, specifically the document layout analysis task, publicly available document layout analysis datasets are utilized, including PubLayNet~\cite{zhong2019publaynet}, DocLayNet~\cite{Pfitzmann_2022}, and Docbank~\cite{li2020docbank}. Data for another region-level task, table understanding, is sourced from PubTabNet~\cite{zhong2020image} with its table annotations.
All data is transformed into the instruction format illustrated in ~\cref{fig:layout_instruct}(a).
In total, 5.7 million instructions are constructed, with a ratio of 1:4:4 for document-level, region-level, and segment-level tasks, respectively.
For detailed instruction templates and dataset descriptions, please refer to the supplementary material.

\noindent{\bf{Layout-aware SFT data}} of LayoutLLM is generated by GPT (GPT-3.5 Turbo) and converted from existing textual Machine Reading Comprehension (MRC) datasets, as discussed in \cref{sec:layoutsft}.
To generate high-quality document-based textual QA and textual CoT, it is essential to make GPT comprehend the document layout.
So, the document is represented using both layout text~\cite{he2023icl} and HTML.
Similar to the pre-training data, the $D_I$ in \cref{alg:LayoutCoT_algorithm} is also randomly sampled from PubLayNet~\cite{zhong2019publaynet}, DocLayNet~\cite{Pfitzmann_2022}, Docbank~\cite{li2020docbank}, RVL-CDIP~\cite{harley2015evaluation}, and DocILE~\cite{2023docile} for building layout text.
The $D_H$ in \cref{alg:LayoutCoT_algorithm} is from GPT's free generation.
The $D_M$ in \cref{alg:LayoutCoT_algorithm} is randomly sampled from the FeTaQA~\cite{nan2022fetaqa} which is a wikipedia question answering dataset.
A total of 300K instructions are constructed, with a ratio of 5:4.5:0.5 for $D_I$, $D_H$, and $D_M$, respectively.
For detailed prompt templates of document-based text QA and text CoT generation using GPT, prompts for HTML generation using GPT, and dataset description, please refer to the supplementary material.

\subsection{Training Setup}
The encoder weight of LayoutLLM is initialized from the LayoutLMv3-large~\cite{huang2022layoutlmv3} which is a widely-used document pre-trained model.
And the LLM backbone is initialized from Vicuna-7B-v1.5~\cite{zheng2023judging}.
Other parameters are randomly initialized.
During pre-training, the LLM is frozen, and the parameters of the two projectors and document pre-trained model encoder are updated.
During SFT, both LLM and two projectors are fine-tuned while keeping the document pre-trained model encoder frozen.
For detailed training setup, please refer to the supplementary material.


\begin{table*}[htp]
    \small 
    \centering
    \vspace{-3mm}
    \begin{tabular}{
          m{.27\columnwidth}<{\centering\arraybackslash}|
          m{.45\columnwidth}|
          m{.17\columnwidth}<{\centering\arraybackslash}|
          m{.23\columnwidth}<{\centering\arraybackslash}|
          m{.17\columnwidth}<{\centering\arraybackslash}|
          m{.17\columnwidth}<{\centering\arraybackslash}|
          m{.17\columnwidth}<{\centering\arraybackslash}
      }
      \toprule[1pt]
      \multirow{2}*{} & 
      \multirow{2}*{\textbf{Models}} & 
      \multicolumn{2}{c|}{\textbf{Document VQA}} & 
      \multicolumn{3}{c}{\textbf{QA for VIE}}\\
      \cline{3-7}
       & & \bf{DocVQA} & \bf{VisualMRC} & \bf{FUNSD} & \bf{CORD} & \bf{SROIE} \\
       \hline
      \multirow{1}{*}{\makecell{\bf{Fine-tuned PTM}}} 
          & LayoutLMv3~\cite{huang2022layoutlmv3} & 83.37$^*$ & - & 92.08$^{*\ddag}$ & 97.46$^{*\ddag}$ & - \\
      \hline
      \multirow{4}{*}{\makecell{\bf{LLM}\\\textit{Plain Text}}} 
          & Llama2-7B~\cite{touvron2023llama2} & 61.34 & 29.73 & 40.78 & 4.39 & 15.86 \\
          & Llama2-7B-chat~\cite{touvron2023llama2} & 64.99 & 52.84 & 48.20 & 47.70 & 68.97 \\
          & Vicuna-7B~\cite{zheng2023judging} & 61.39 & 53.63 & 49.79 & 44.67 & 67.49 \\
          & Vicuna-1.5-7B~\cite{zheng2023judging} & 66.99 & 52.13 & 48.06 & 51.40 & 68.20 \\ 
      \hline
      \multirow{4}{*}{\makecell{\bf{LLM}\\\textit{Layout Text}\\(Text + Box)~\cite{he2023icl}}}
          & Llama2-7B~\cite{touvron2023llama2} & 37.32 & 33.82 & 51.40 & 28.04 & 34.96 \\
          & Llama2-7B-chat~\cite{touvron2023llama2} & 56.55 & 49.26 & 58.34 & 50.93 & 51.15 \\
          & Vicuna-7B~\cite{zheng2023judging} & 37.21 & 52.55 & 42.73 & 46.59 & 45.43 \\
          & Vicuna-1.5-7B~\cite{zheng2023judging} & 56.81 & 47.22 & 59.63 & 56.13 & 66.20 \\ 
      \hline
      \multirow{5}{*}{\textbf{MLLM}} 
          & LLaVAR-7B~\cite{zhang2023llavar} & 11.6\dag & 36.37 & 1.71 & 13.55 & 2.38 \\
          & LLaVA-1.5-7B~\cite{liu2023improvedllava} & 13.34 & 35.23 & 1.93 & 18.06 & 3.83 \\
          & mPLUG-DocOWL-7B~\cite{ye2023mplugdoc} & 62.2$^{*}$\dag & - & - & - & - \\
          & Qwen-VL-7B~\cite{bai2023qwen} & 65.1$^{*}$\dag & 42.52 & 47.09 & 30.00 & 58.59 \\
      \hline
        & \bf{LayoutLLM-7B$\triangle$ (Ours)} & \bf{74.25} & \bf{55.73} & \bf{78.65} & \bf{62.21} & \bf{70.97}  \\
        & \bf{LayoutLLM-7B$\star$ (Ours)} & \bf{74.27} & \bf{55.76} & \bf{79.98} & \bf{63.10} & \bf{72.12}  \\
      \bottomrule[1pt]
    \end{tabular}
    \vspace{-2mm}
    \caption{
      Zero-shot document understanding results on open-source LLMs and MLLMs.
      $*$ signifies training set use; unmarked results are zero-shot.
      Results marked with $^{\ddag}$ are F1 scores for VIE.
      Results marked with $\dag$ are from the original paper and others are re-implemented by us.
      $\triangle$ marks LayoutLLM's LLM backbone as initialized with Llama2-7B-chat, and $\star$ with Vicuna-1.5-7B.
    }
    \label{table:main_results}
    \vspace{-4mm}
\end{table*}

\subsection{Evaluation Setup}
\label{sec:eval_setup}
The zero-shot ability is highly expected in real-world document understanding scenarios~\cite{cui2021document,liu2023hidden,shi2023exploring}.
Therefore, zero-shot evaluations on widely-used document understanding benchmarks including document visual question answering (Document VQA) and visual information extraction (VIE) are conducted.
Only the test sets are utilized in all benchmarks and only the official provided image, text, and layout information are used.
The Document VQA datasets comprise the \textbf{DocVQA}\cite{mathew2021docvqa} test set, consisting of 5,188 questions, and the \textbf{VisualMRC}\cite{tanaka2021visualmrc} test set containing 6,708 questions.
Following the evaluation metric settings of the original datasets, the ANLS~\cite{mathew2021docvqa} is utilized for evaluating DocVQA, and Rouge-L is used for evaluating VisualMRC.
For the VIE task, \textbf{FUNSD}~\cite{jaume2019funsd}, \textbf{CORD}~\cite{park2019cord}, and \textbf{SROIE}~\cite{Huang_2019} are used.
The test set of FUNSD comprises 50 form images, each annotated with entity-level headers, questions, answers, and others, along with entity linking annotations.
CORD's test set consists of 100 receipt images, annotated with 30 entity types, such as the tax amount, total price, etc.
SROIE's test set includes 347 receipt images, annotated with 4 entity types: company, date, address, and total.
To prompt LLMs/MLLMs for zero-shot VIE, annotations in VIE datasets are transformed into question answering format (QA for VIE).
For key-value annotations with linking in the FUNSD, the format is \{Q: What is the ``key" in the document? A: ``value"\}.
For entity annotations in CORD and SROIE, directly asking for the entity in the document like \{Q: What is the address in the document? A: ``the address annotation"\} is utilized.
Following DocVQA, the QA for VIE task is evaluated by ANLS.



\subsection{Main Results}

As shown in \cref{table:main_results}, the zero-shot document understanding performance of LayoutLLM and existing open-source LLMs and MLLMs is evaluated.
Generally, the existing LLMs are better than MLLMs for zero-shot document VQA and VIE.
For example in the results on DocVQA, most LLMs can achieve a performance of around 60\% or higher, while most MLLMs can only attain around 10\%, except for mPLUG-DocOWL and Qwen-VL that trained with the training set.
One possible reason is that it's difficult for these MLLMs to obtain accurate textual information from document images.
Additionally, for LLMs, using Plain Text and Layout Text respectively as the document representation are further discussed, where the Layout Text introduces layout information by adding text coordinates in the format: \{text:``text", box:[x1,y1,x2,y2]\}~\cite{he2023icl}.
Compared to the Plain Text, the Layout Text variant doesn't show stable performance improvements, noticeable in certain tasks, for example in Vicuna-1.5, an improvement in VIE (FUNSD 48.06\% to 59.63\%) but a decline in DocVQA (66.99\% to 56.81\%).
LLMs may lack the ability to learn this formatted layout text, and directly adding layout information (e.g., coordinates) to the text will also highly increase the token length, making the answer inference more challenging.

Compared to the prior SOTA model, LayoutLMv3, which is fine-tuned using the training set of downstream tasks, LayoutLLM demonstrates competitive performance on the DocVQA benchmark.
Compared with these LLMs and MLLMs, LayoutLLM achieves consistent and significant improvements over them on all evaluation benchmarks.
Notably, LayoutLLM which employs zero-shot performance, outperforms mPLUG-DocOWL and Qwen-VL by around 10\% on the DocVQA dataset, both of which are trained with this dataset. 
This demonstrates that LayoutLLM can learn more robust and discriminative representations for document understanding.
Furthermore, experiments of the different initialization of the LLM backbone have all achieved optimal results across all benchmarks, substantiating that LayoutLLM can adapt to various LLMs.
In summary, our method explores a more effective way to utilize layout information for document understanding, which significantly improves the performance of zero-shot document understanding.


\subsection{Ablation study}

\begin{table}[t!]
    \footnotesize
    \vspace{-2mm}
    \begin{center}
        \resizebox{1\linewidth}{!}{
            \begin{tabular}{
                m{.05\columnwidth}<{\centering}|
                m{.23\columnwidth}<{\centering}
                m{.23\columnwidth}<{\centering}|
                m{.16\columnwidth}<{\centering}|
                m{.12\columnwidth}<{\centering}
            }
            \toprule[1pt]
            \# & \makecell{Layout-aware\\Pre-training} & \makecell{Layout-aware\\SFT} & DocVQA & FUNSD \\
            \hline
            0 & & & 70.82 & 70.96  \\
            1 & \checkmark &  & 72.31 & 74.02 \\
            2 & \checkmark & \checkmark & \bf{74.27} & \bf{79.98} \\
            \bottomrule[1pt]
            \end{tabular}
        }
    \vspace{-2mm}
    \caption{Ablation study on the DocVQA and FUNSD test sets.}
    \vspace{-11mm}
    \label{table:ablation_results}
    \end{center}
\end{table}

To better verify the effectiveness of the layout-aware pre-training and layout-aware SFT in the layout instruction tuning, an ablation study is conducted (see \cref{table:ablation_results}).

\noindent\textbf{Initial baseline.} The \#0 baseline disables both layout-aware pre-training and layout-aware SFT.
It only adopts SFT (same SFT data but without LayoutCoT steps) on the LayoutLLM.
Even without any alignment pre-training and LayoutCoT steps guidance, the baseline outperforms previous SOTAs, achieving 70.82\% on DocVQA and 70.96\% on FUNSD.
This indicates the document understanding ability of DocPTM benefits document understanding with LLMs.
\textbf{Effect of Layout-aware pre-training.} In \#1, the layout-aware pre-training and the same SFT with \#0 is conducted.
Compared to \#0, benefiting from the basic document understanding capability learned through the layout-aware pre-training, \#1 shows an increase of 1.49\% on DocVQA and 3.06\% on FUNSD.
It can be observed that the basic document understanding ability learned from the layout-aware pre-training significantly enhances the performance of the basic key-value extraction tasks in the FUNSD dataset.
Compared to the FUNSD VIE task, it also shows that the DocVQA is a more complex task.
\textbf{Effect of Layout-aware SFT.} Compared with \#1, \#2 further incorporates the layout-aware SFT strategy, resulting in a performance gain of 1.96\% on DocVQA and even 5.96\% on FUNSD. This indicates the LayoutCoT in layout-aware SFT can help the LayoutLLM to handle complex document tasks and it is effective for boosting the performance on both document VQA and VIE. Moreover, it also promotes a certain degree of interpretability. Overall, the ablation study demonstrates the importance of layout-aware pre-training and layout-aware SFT for zero-shot document understanding.

\subsection{Qualitative Results}

\begin{figure}[tp]\centering
    \vspace{-3mm}
    \setlength{\abovecaptionskip}{0mm}
     \includegraphics[width=0.70\linewidth]{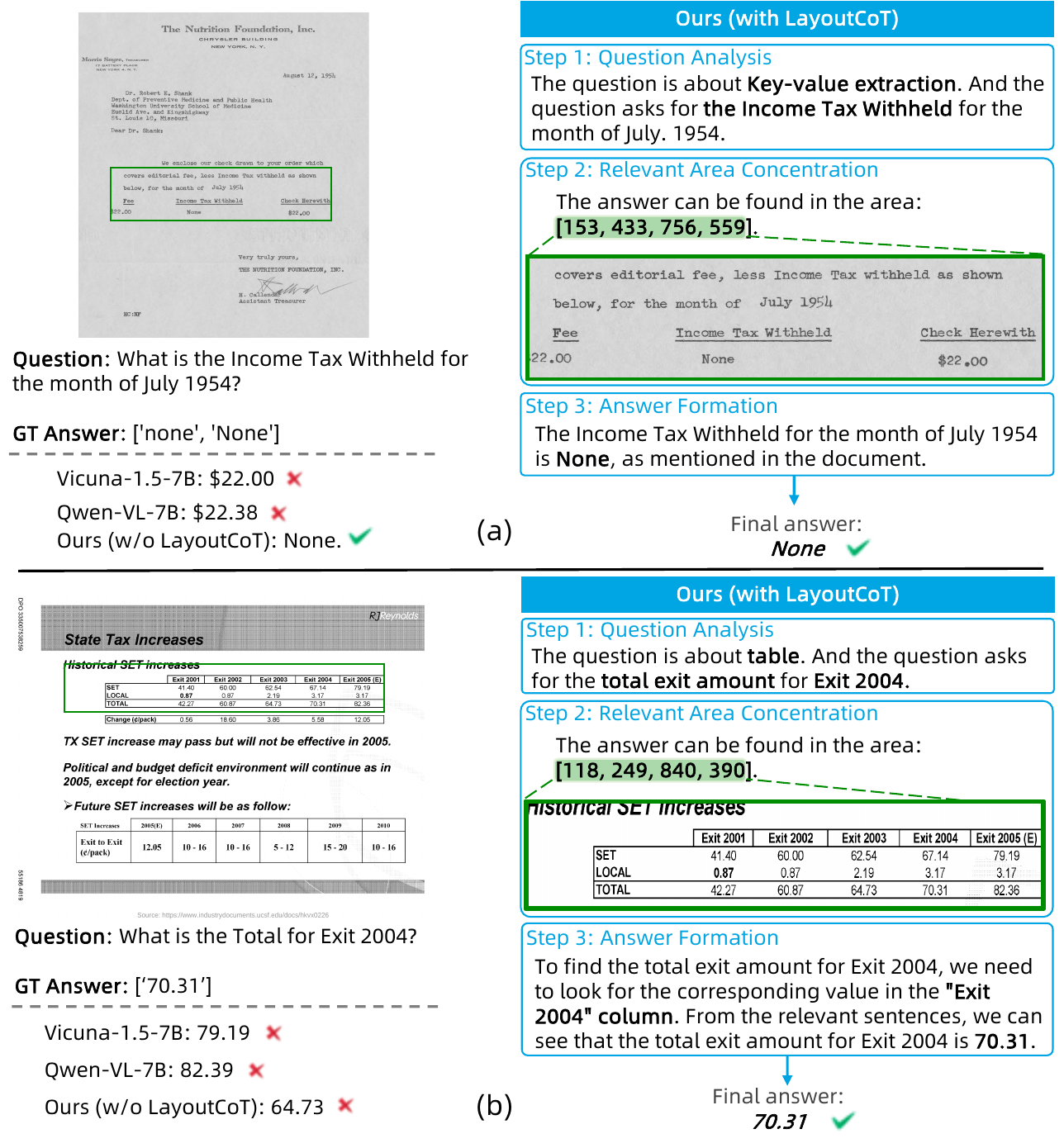}
    \caption{Qualitative results on DocVQA. \textcolor{Green}{\textbf{Green}} boxes are the areas concentrated in the step 2 of LayoutCoT.}
    \vspace{-6mm}  
     \label{fig:visualization}
\end{figure}

Two examples are shown in \cref{fig:visualization}. Through combined with layout-aware pre-training and layout-aware SFT, LayoutLLM can accurately focus on the relevant areas, utilize the layout information to assist in problem-solving and provide interpretability. For example, in \cref{fig:visualization}(a) question about key-value extraction in up-down layout, different from the left-right variant, relies more on document layout to infer the right answer.
Since the keywords ``Income Tax'' in the question often co-occur with numerical data, Vicuna-1.5 and Qwen-VL find numerical answers relying more on the semantics than the layout, resulting in incorrect responses.
In contrast, benefiting from the layout-aware pre-training, our model can effectively leverage layout information to give accurate answers.
In addition, the model using LayoutCoT can further provide the location and the reasoning process, showing a certain degree of interpretability.
But in certain situations, only combined with layout pre-training, our model might fail to give accurate answers.
As shown in \cref{fig:visualization}(b), without LayoutCoT, our model identifies ``Exit 2003" as the relevant column and generates a wrong answer. 
However, with the help of LayoutCoT, LayoutLLM can correctly identify the question type as ``Table", locate the relevant table area, and finally infer the right answer from the corresponding ``Exit 2004" column.

\begin{figure}[tp]\centering·

    \vspace{-3mm}
    \includegraphics[width=0.75\linewidth]{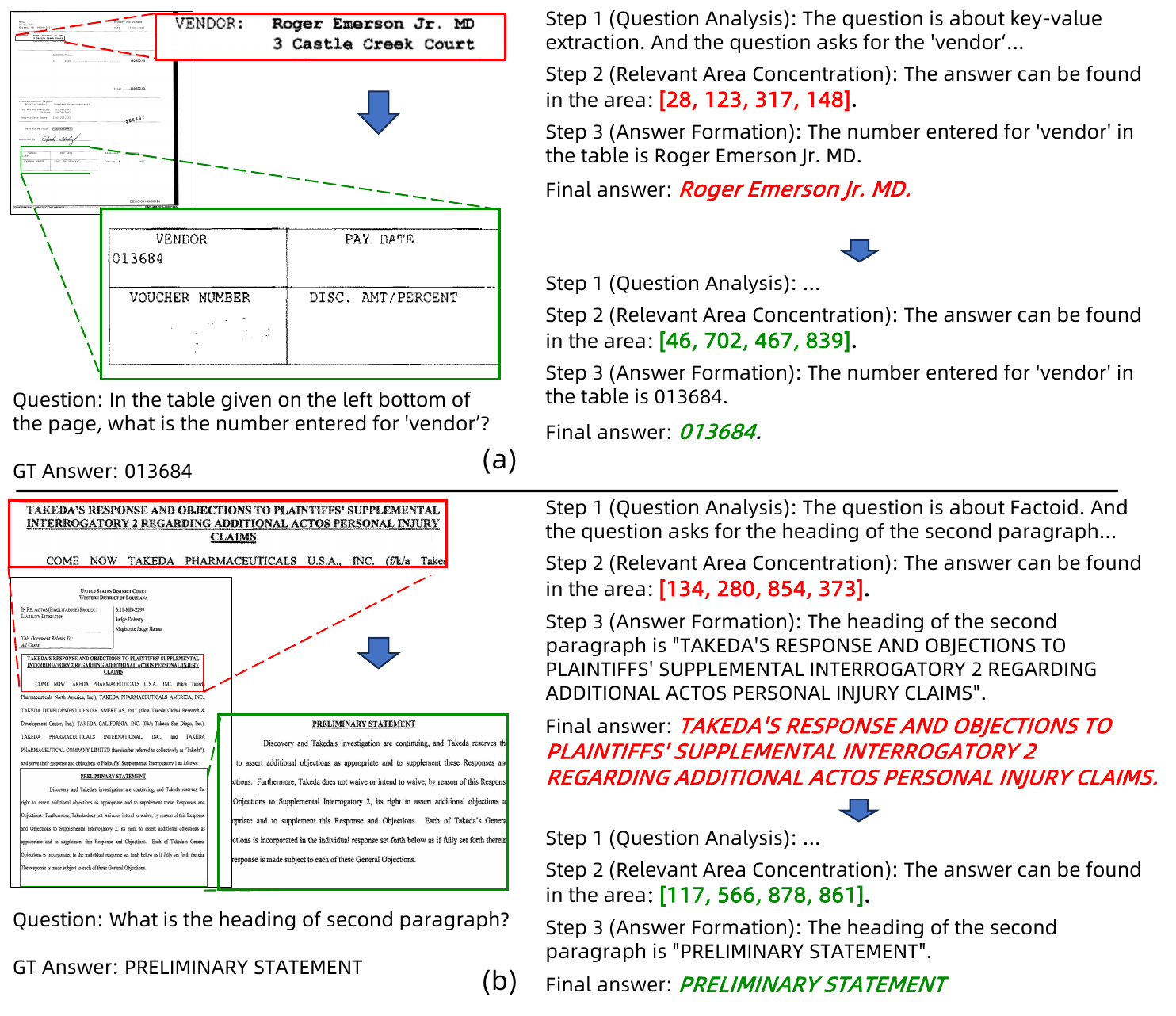}
    \setlength{\abovecaptionskip}{1mm}
    \vspace{-1mm}
    \caption{Interactive correction with LayoutCoT. 
    \textcolor{Green}{\textbf{Green}} represents the correct relevant areas and answers, while \textcolor{red}{\textbf{Red}} represents the original incorrect ones. Best viewed in digital version.}
    \label{fig:visualization_editing}
    \vspace{-4mm}
\end{figure}

\subsection{Interactive Correction with LayoutCoT} \label{sec:interactive}

Since LayoutCoT runs in a step-by-step fashion and produces intermediate results in the inference stage, it can facilitate \textit{interactive inspection and correction}, when processing a document. As shown in Fig.~\ref{fig:visualization_editing}(a), there are two areas in the image that are relevant to the keyword ``vendor" of the question. LayoutCoT focused on a wrong area containing ``vendor", as it missed ``left bottom" in the question, the answer was incorrect. However, after the right area is given manually, it can finally give the correct answer. Similarly, in Fig. \ref{fig:visualization_editing}(b) the question asks about ``the heading of the second paragraph". However, the term ``paragraph" does not have a universal definition, and in this case, the sentences in the area below the main heading were considered as a \textit{paragraph}, causing the model to predict ``TAKEDA's...CLAIMS", which was incorrect according to the GT. Once the right ``second paragraph" region is fed to the model, the answer can be successfully revised. This unique ability of LayoutCoT could be very valuable in high-stake scenarios (e.g., a bank transaction), where the standards are extremely high, and manual checking and correction (i.e., human-in-the-loop) are required.

\section{Limitations}

Through LayoutCoT, LayoutLLM demonstrates the capability of interactive correction, but in real-world applications, this is not enough. The ability to refuse false-positive outputs and generate hints (e.g. ``The answer is not mentioned in the document.") is crucial. However, it is currently absent in LayoutLLM. In addition, despite achieving notable improvements through layout-aware pre-training, LayoutLLM struggles in precisely understanding region-level relationships, as evidenced in \cref{fig:visualization_editing}(a). We will study how to endow LayoutLLM with such abilities.


\section{Conclusion}

We propose LayoutLLM for document understanding, in which a layout instruction tuning strategy comprising layout-aware pre-training and layout-aware SFT is designed to improve the comprehension of document layouts. Extensive experiments confirm the effectiveness of LayoutLLM.
{
    \small
    \bibliographystyle{ieeenat_fullname}
    \bibliography{main}
}

\clearpage
\setcounter{page}{1}
\maketitlesupplementary
\renewcommand\thesection{\Alph{section}}
\setcounter{section}{0}

\section{Dataset Details}


\subsection{Dataset Description}
Details of the publicly available document understanding datasets used in layout-aware pre-training and layout-aware SFT are as follows:

\noindent{\textbf{RVL-CDIP}}~\cite{harley2015evaluation} is a document classification dataset that consists of 400,000 grayscale document images in 16 classes. The 16 classes encompass diverse document types such as letter, form, email, handwritten, advertisement, scientific report, scientific publication, specification, file folder, news article, budget, invoice, presentation, questionnaire, resume, and memo.
To enhance the diversity of the data in training, sampling is performed within each class.
The text and layout information are extracted by Tesseract OCR.

\noindent{\textbf{DocILE}}~\cite{2023docile} is a large dataset for key information localization and extraction.
It consists of 6,680 annotated business documents.
And there are 932k unlabeled documents and 100k synthetically generated documents in the dataset for unsupervised pre-training.
It also provides both word-level text and corresponding locations for the document images in the dataset by using DocTR OCR.

\noindent{\textbf{PubLayNet}}~\cite{zhong2019publaynet} is a document layout analysis dataset that covers typical document layout elements such as text, title, list, figure, and table.
It consists of over 360K PDF documents sourced from articles on PubMed Central. 
By parsing the PDF page using PDFMiner and matching the layout with the XML representation, annotations for both page layout and text information are generated.

\noindent{\textbf{DocBank}}~\cite{li2020docbank} is a document layout analysis dataset that contains 500K document pages with fine-grained token-level annotations for document layout analysis.
These documents are sourced from articles on arXiv.org, covering various areas such as Physics, Mathematics, Computer Science, and others.
PDFPlumber, which is a PDF parser built on PDFMiner, is used to extract text lines and non-text elements with their bounding boxes.

\noindent{\textbf{DocLayNet}}~\cite{Pfitzmann_2022} is a human-annotated document layout segmentation dataset containing 80863 manually annotated PDF documents with 11 complex layout classes from diverse data sources including Finance, Science, Patents, Tenders, Law texts, and Manuals.
DocLayNet also provides original PDF pages, parsed PDF text, and text-cell coordinates.

\noindent{\textbf{PubTabNet}}~\cite{zhong2020image} is a large-scale table recognition dataset containing over 568K tables from scientific publications.
It contains heterogeneous tables in both image and HTML format.
To support more diverse model designs, it also provides the position (bounding box) of table cells.

\noindent{\textbf{FeTaQA}~\cite{nan2022fetaqa}} is a dataset consisting of 10K Wikipedia-based table question answering pairs along with tabular data extracted from webpages. These question answering pairs require complex reasoning and integration of table information. 
The tabular data supplied can be rendered into an HTML document, facilitating document understanding tasks.
FetaQA also provides annotations of highlighted cells, which are table regions corresponding to question answering pairs, and these annotations are used to construct the intermediate inference information for LayoutCoT.

\subsection{Layout-aware Pre-training Data Construction}
This section describes the data construction process for layout-aware pre-training tasks used to train LayoutLLM, including the preprocessing of document data, the prompt templates for prompting GPT (GPT-3.5 Turbo), and the instruction templates utilized during training.

\begin{figure*}[tp]
    \centering
    \includegraphics[width=.99\linewidth]{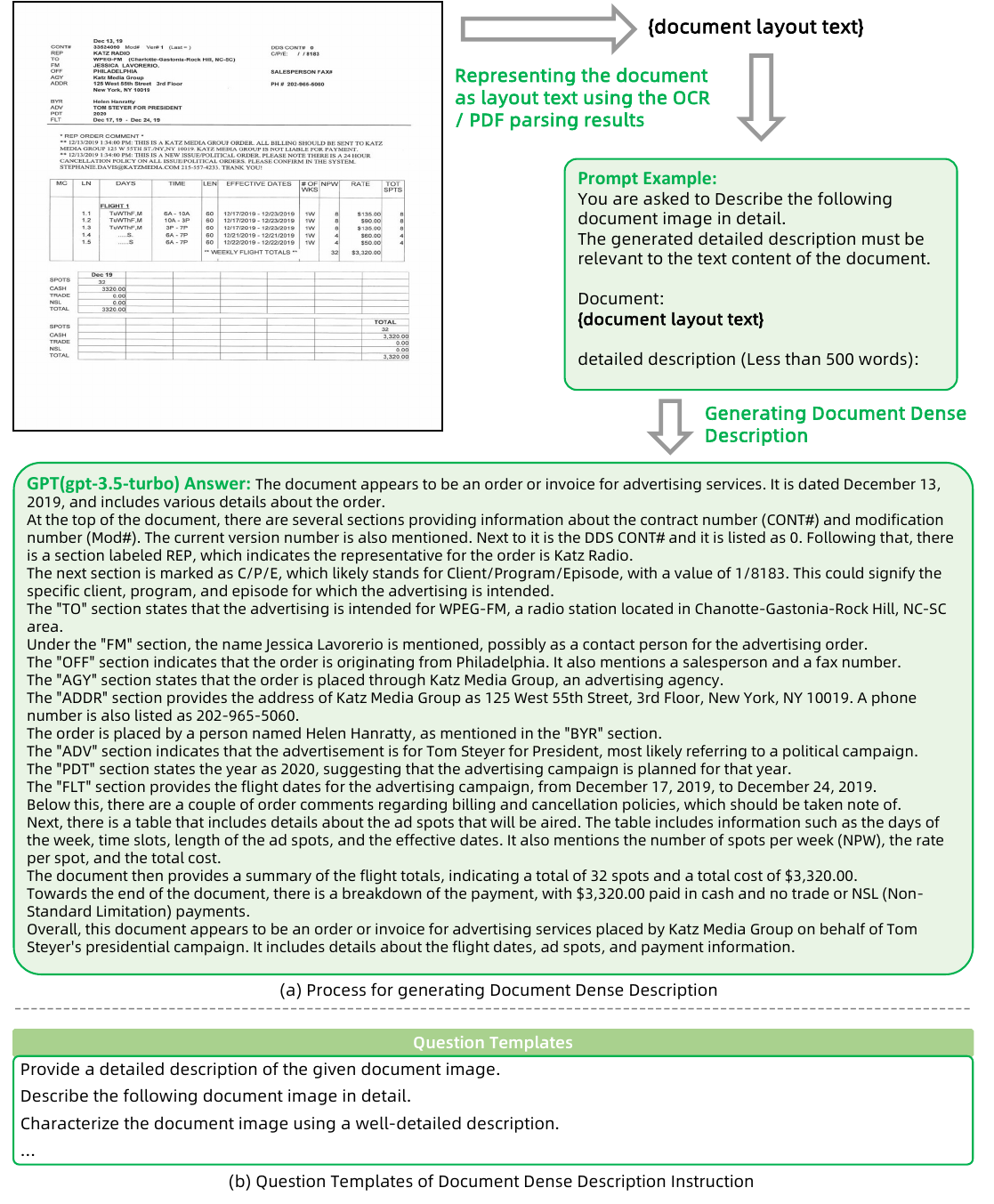}
    \caption{The data construction of Document Dense Description (DDD) task for document-level pre-training. (a) Given a document, generating its dense description as the instruction answer using its layout text with the help of GPT; (b) The shown question templates are randomly sampled as DDD instructions.} 
    \label{fig:doc_level_ddd}
    \vspace{-4mm}
\end{figure*}

\begin{figure*}[tp]
    \centering
    \includegraphics[width=.99\linewidth]{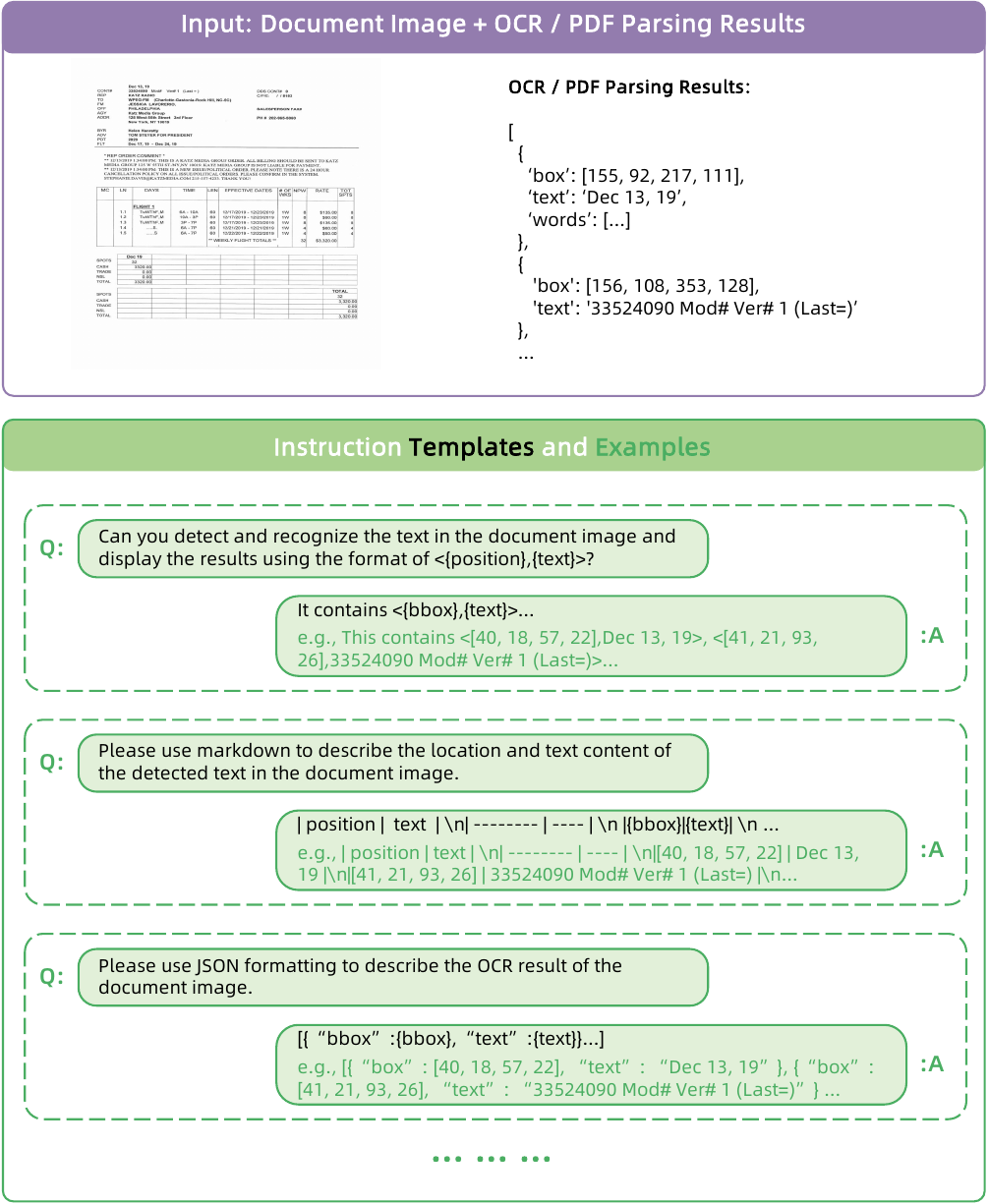}
    \caption{The data construction of Text and Layout Reconstruction (TLR) task for document-level pre-training. Given a document image with OCR or PDF parsing results, constructing TLR instructions by reconstructing the complete text and layout information of the document in a specific format.}
    \label{fig:doc_level_ocr}
    \vspace{-4mm}
\end{figure*}

As mentioned in \cref{sec:data_collection}, GPT (GPT-3.5 Turbo) is used to generate detailed descriptions of documents for the document-level pre-training task, Document Dense Description (DDD).
\cref{fig:doc_level_ddd} illustrates the data generation process for the DDD task and the instruction templates for pre-training DDD.
As \cref{fig:doc_level_ddd} (a) shows, the document is represented as layout text.
Considering training efficiency and to prevent excessively long generated content, a requirement that the number of words generated should be less than 500 is in the prompt.
During pre-training, the instructions that are shown in \cref{fig:doc_level_ddd} (b) are randomly sampled as question instructions for DDD.
\cref{fig:doc_level_ocr} shows the instruction templates for the document-level pre-training task: the Text and Layout Reconstruction (TLR).
The TLR task aims to reconstruct the complete text and layout information of a document and present it in a specific format, such as the prescribed ``\textless \{box\}, \{text\} \textgreater" format, JSON format, or Markdown format.

\begin{figure*}[tp]
    \centering
    \includegraphics[width=.87\linewidth]{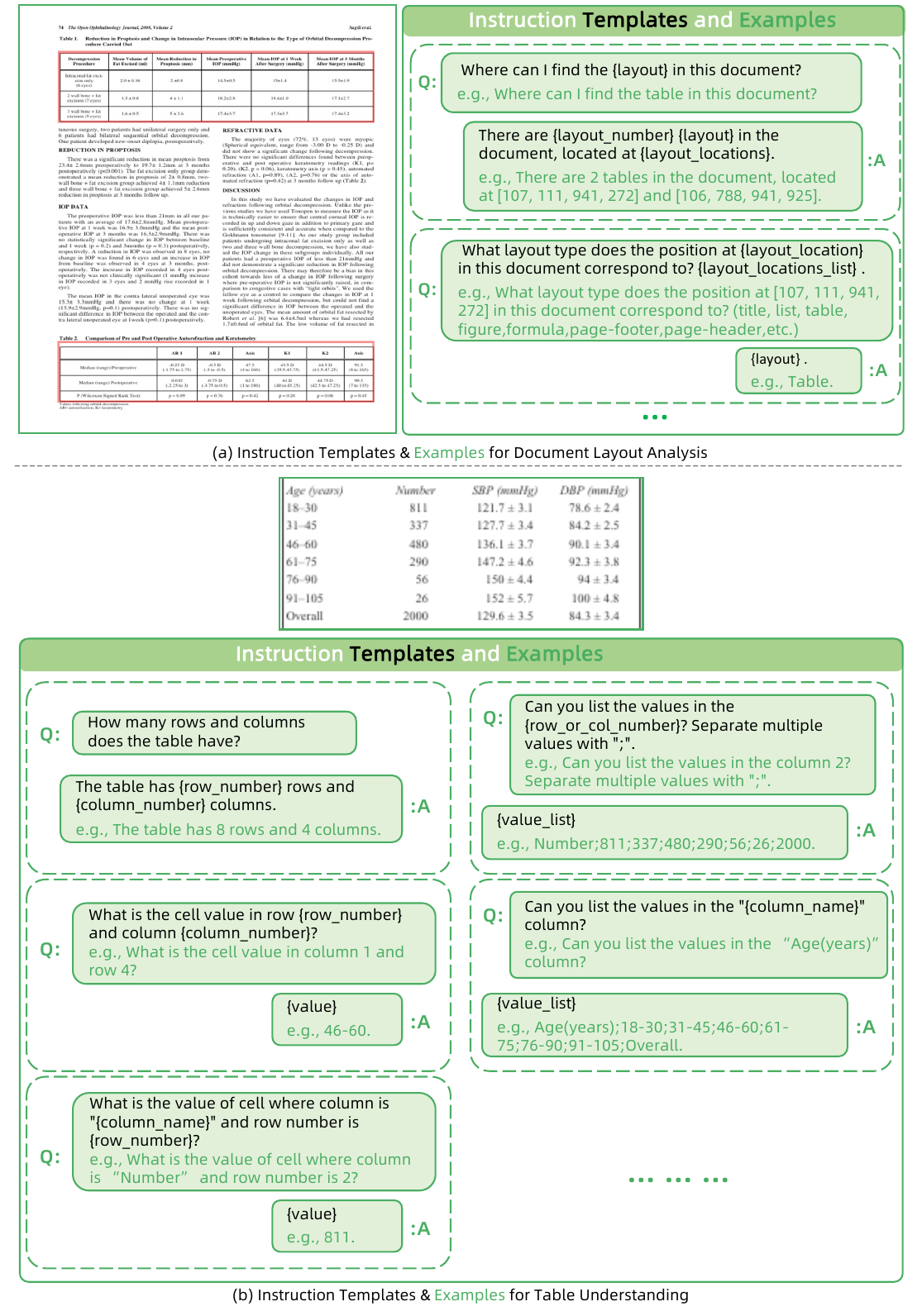}
    \caption{The data construction of Document Layout Analysis task and Table Understanding task for region-level pre-training. (a) Constructing Document Layout Analysis task instructions by transforming layout annotations to the presented templates; (b) Constructing Table Understanding task instructions by transforming table annotations to the presented templates.}
    \label{fig:regin_level}
\end{figure*}

For region-level pre-training tasks, pre-training is conducted by transforming document layout annotations and table annotations from the original dataset into instructions, as illustrated in the templates shown in \cref{fig:regin_level}.

\begin{figure*}[tp]
    \centering
    \includegraphics[width=.85\linewidth]{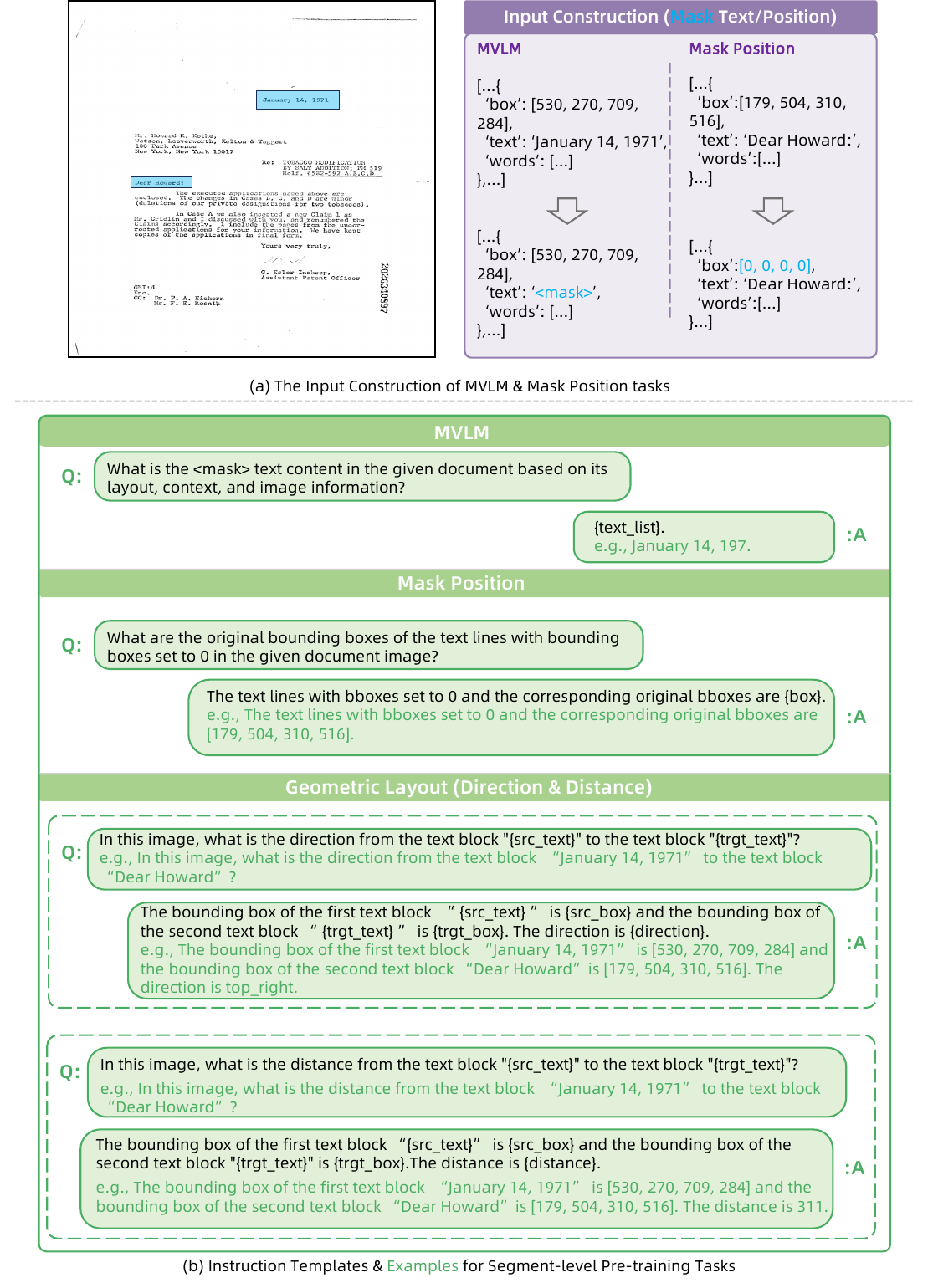}
    \caption{The data construction of MVLM instruction, Mask Position instruction and Geometric Layout instruction tasks for segment-level self-supervised pre-training. (a) For the MVLM and Mask Position instructions, randomly \textcolor[RGB]{0,167,238}{masking} the text or position coordinates respectively as the inputs for LayoutLLM's document pre-trained model encoder. The masked information is utilized as the target for self-supervised learning. (b) For the three segment-level pre-training tasks, constructing the instructions for text masking, position masking, and geometric-related tasks based on the presented templates respectively.}
    \label{fig:seg_level}
    \vspace{-4mm}
\end{figure*}

The segment-level pre-training tasks are all self-supervised.
Utilizing the original text and layout information, the input involves masking text or coordinates, or performing geometric-related calculations based on coordinates to obtain corresponding self-supervised targets.
The obtained self-supervised targets are transformed into instructions in \cref{fig:seg_level} for pre-training.


\subsection{Layout-aware SFT Data Construction}
\begin{figure*}[tb]
    \centering
    \includegraphics[width=.9\linewidth]{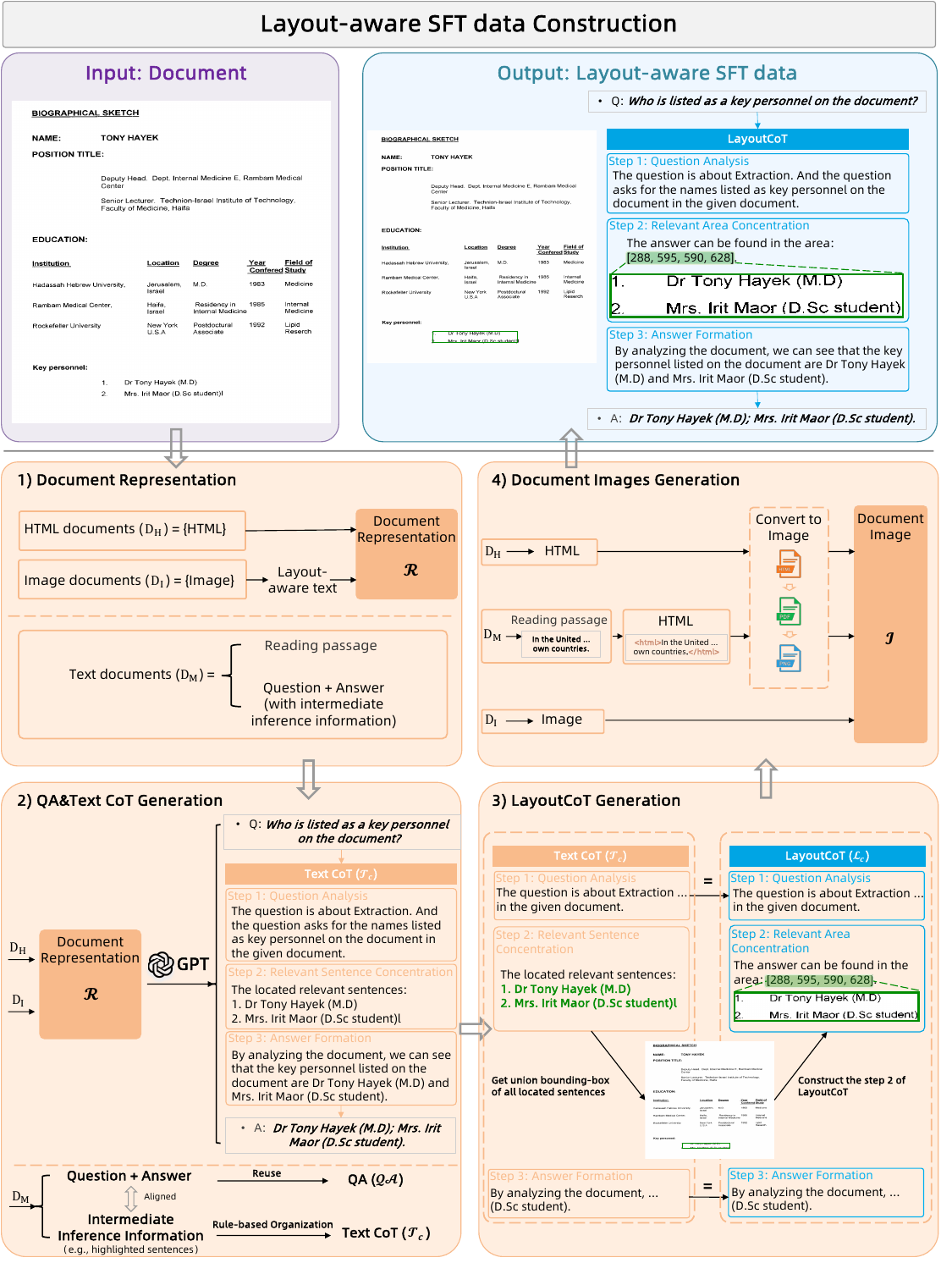}
    \caption{Overview of Layout-aware SFT data construction, involving 4 steps: Document Representation, QA \& Text CoT Generation, LayoutCoT Generation, and Document Images Generation.}
    \label{fig:app_sft_overall}
    \vspace{-4mm}
\end{figure*}

This section describes the process of constructing layout-aware SFT data. \cref{fig:app_sft_overall} presents an overall construction process of the layout-aware SFT data from \cref{alg:LayoutCoT_algorithm}, including the document representation, QA\&Text CoT Generation, LayoutCoT Generation, and Document Images Generation.

\begin{figure*}[tb]
    \centering
    \includegraphics[width=.9\linewidth]{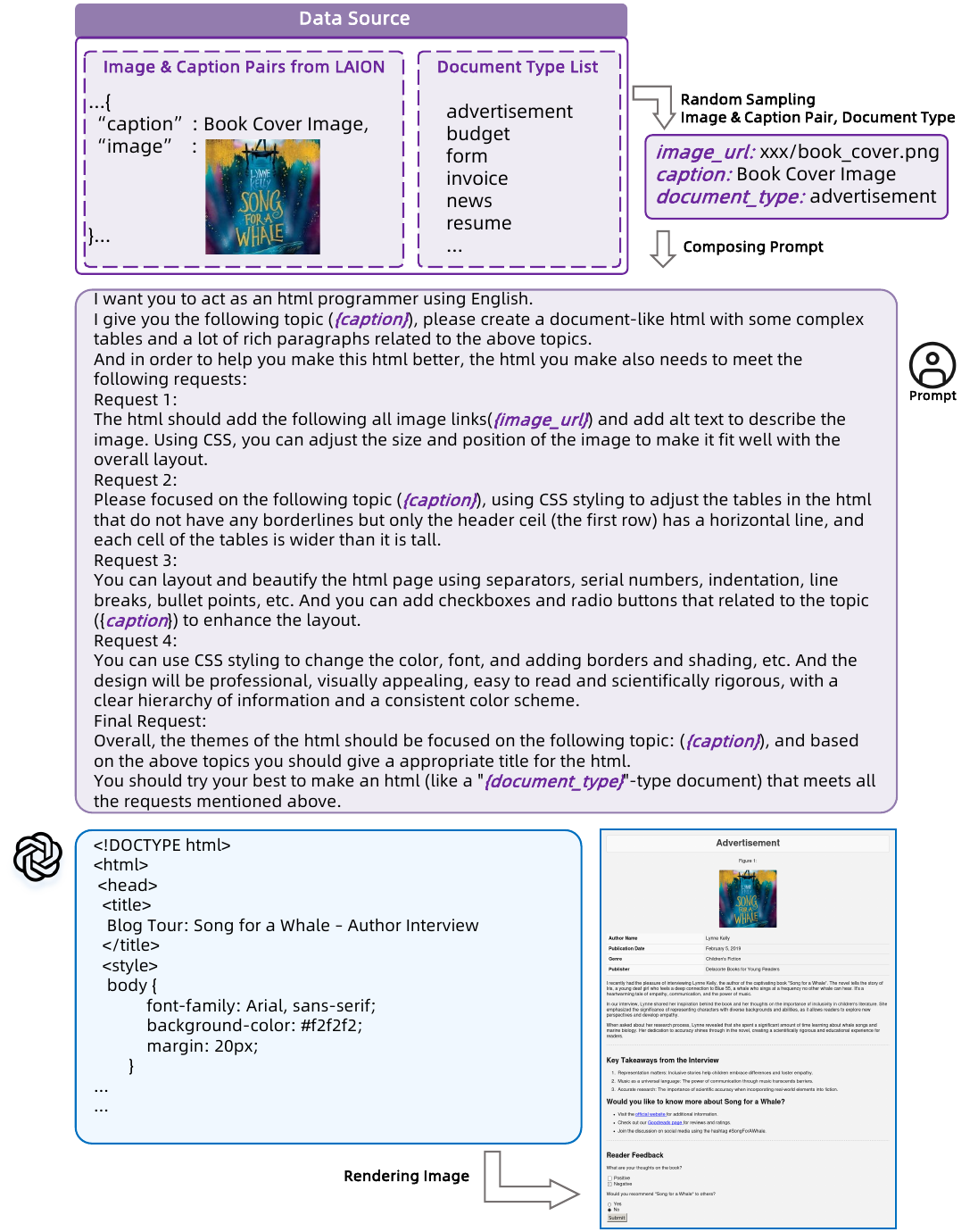}
    \caption{An example of the HTML documents $D_H$ generation pipeline using GPT. To generate HTML documents with diverse layout types by GPT, the data source (including image \& caption pairs from LAION and document type list) are randomly sampled as the inputs for composing the prompts of $D_H$ generation.}
    \label{fig:app_sft_html}
    \vspace{-4mm}
\end{figure*}

\begin{figure*}[tp]
    \centering
    \includegraphics[width=.78\linewidth]{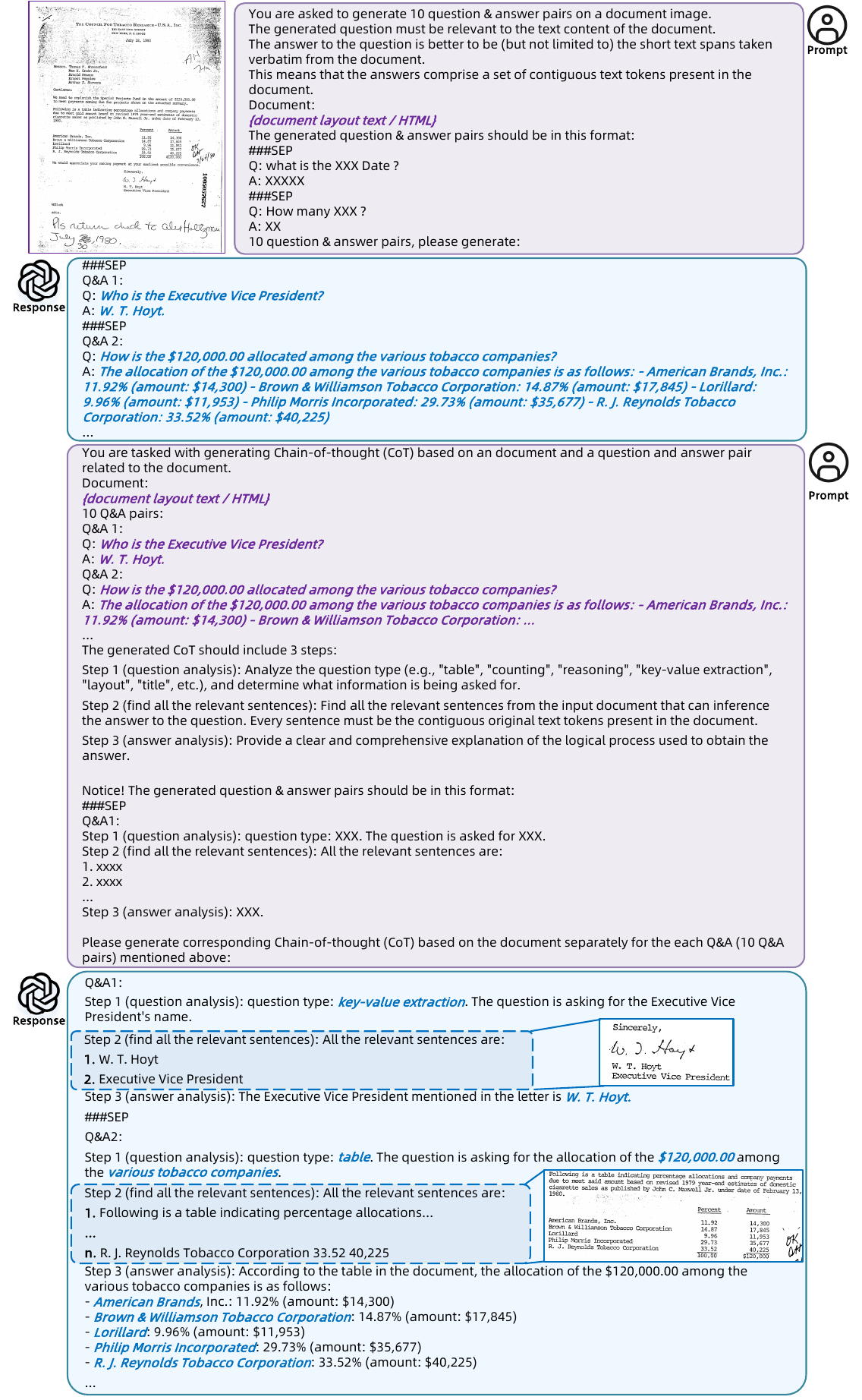}
    \caption{An example of QA and corresponding Text CoT ($\mathcal{T}_c$) generation. Given a document representation (layout text or HTML), generating its QA \& Text CoT with the help of GPT.}
    \label{fig:sft_doc_qa}
    \vspace{-4mm}
\end{figure*}

For the document representation, image documents ($D_I$) and text documents ($D_M$) are from publicly available document understanding datasets.
As mentioned in \cref{sec:data_collection}, the HTML document ($D_H$ in \cref{alg:LayoutCoT_algorithm}) is from GPT's free generation. 
\cref{fig:app_sft_html} shows an example of the GPT's free generation pipeline for $D_H$.
To generate diverse HTML documents through GPT, some images and captions are randomly sampled from the LAION-5B dataset, which is a large-scale dataset consisting of more than 5B image-text pairs, as the topic and figure source of the generated $D_H$.
A list of document types is provided for randomly sampling a document type as a constraint when prompting GPT.
The detailed prompt is illustrated in \cref{fig:app_sft_html}.
For the QA\&Text CoT Generation, \cref{fig:sft_doc_qa} shows the prompts and a specific example used for generating $\mathcal{QA}$ and the corresponding Text CoT ($\mathcal{T}_c$) based on document representation ($\mathcal{R}$) using GPT (GPT-3.5 Turbo).
In addition, the $\mathcal{D}_M$ includes the QA pairs and intermediate inference infoamtion process, thereby directly reusing the $\mathcal{QA}$ and organizing $\mathcal{T}_c$ using a rule-based method.
Then, as shown in the \cref{fig:app_sft_overall} (3), the step 1 \& 3 in $\mathcal{T}_c$ are used as the step 1 \& 3 in $\mathcal{L}_c$.
To construct the step 2 (relevant area concentration) of $\mathcal{L}_c$, the union bounding-box of all located relevant sentences in $\mathcal{T}_c$ are taken as the relevant area.
Finally, the text documents ($D_M$) and generated HTML documents ($D_H$) are converted to images.



\section{Training Setup}
The encoder weight of LayoutLLM is initialized from the LayoutLMv3-large~\cite{huang2022layoutlmv3} which is a widely-used document pre-trained model.
And the LLM backbone is initialized from Vicuna-7B-v1.5~\cite{zheng2023judging}.
To further validate the adaptability of the LayoutLLM to different LLMs, llama2-7B-chat~\cite{touvron2023llama2} is also employed, initializing the LayoutLLM's LLM backbone for experiments.
Other parameters are randomly initialized.
$d_0$ is 1024 and $d_1$ is 4096.
Following LayoutLMv3~\cite{huang2022layoutlmv3}, the maximum tokens of the document are set to 512 for training.
The maximum instruction tokens are set to 2048 for training.
The AdamW optimizer is used for both pre-training and SFT.

During pre-training, the LLM is frozen, and the parameters of the two projectors and document pre-trained model encoder are updated.
Only 1 epoch is pre-trained.
The learning rate is set to 1e-4 with a cosine scheduler.
The weight decay is set to 0.0001.
The warmup ratio is 0.03.
The max grad norm is set to 1.0.
The batch size is 32 for each GPU.

During SFT, both LLM and two projectors are fine-tuned while keeping the document pre-trained model encoder frozen.
The SFT stage contains 3 epochs.
The learning rate is set to 2e-5 with a cosine scheduler.
The weight decay is set to 0.
The warmup ratio is 0.03.
The max grad norm is set to 1.0.
The batch size is 8 for each GPU.

\section{Evaluation Setup}
This section introduces the construction of test data for visual information extraction using a QA-based approach.
To prompt LLMs/MLLMs for zero-shot VIE, annotations in VIE datasets are transformed into question answering format (QA for VIE). 

\begin{figure*}[tp]
    \centering
    \includegraphics[width=.55\linewidth]{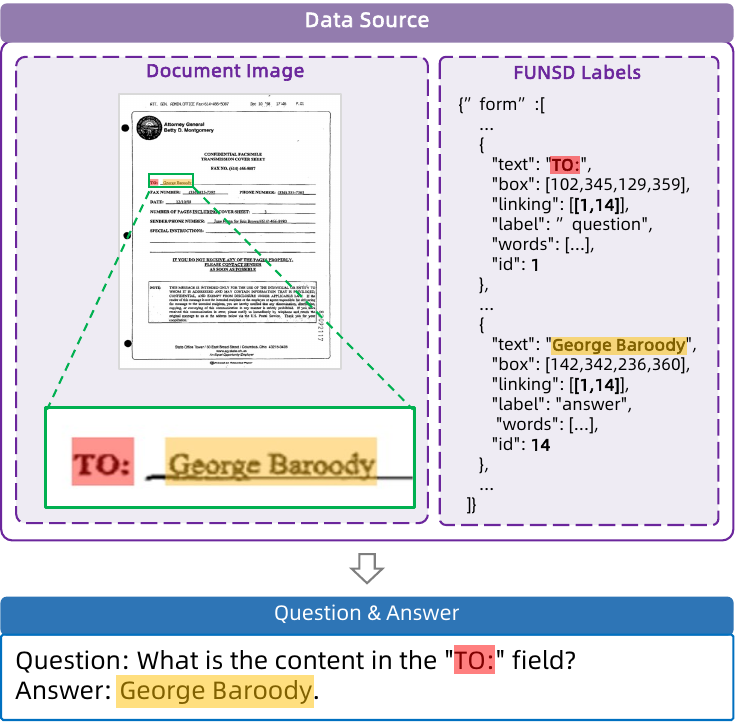}
    \caption{Evaluation data construction example of the QA for VIE task through question-answer with linking annotations in VIE (FUNSD).}
    \label{fig:app_vie_funsd}
    \vspace{-4mm}
\end{figure*}

As mentioned in ~\cref{sec:eval_setup}, the \textbf{FUNSD}~\cite{jaume2019funsd} dataset is a widely used VIE dataset that is annotated in entity-level headers, questions, answers, and others, along with entity linking annotations.
As shown in Fig.~\ref{fig:app_vie_funsd}, for question-answer annotations with linking in the FUNSD, the question answering pairs of evaluation are constructed by asking the values of the linking keys.
To ensure an absolute answer of an evaluation question to avoid ambiguity, cases where one entity links to multiple entities in the FUNSD dataset are filtered out.

\begin{figure*}[tb]
    \centering
    \includegraphics[width=.55\linewidth]{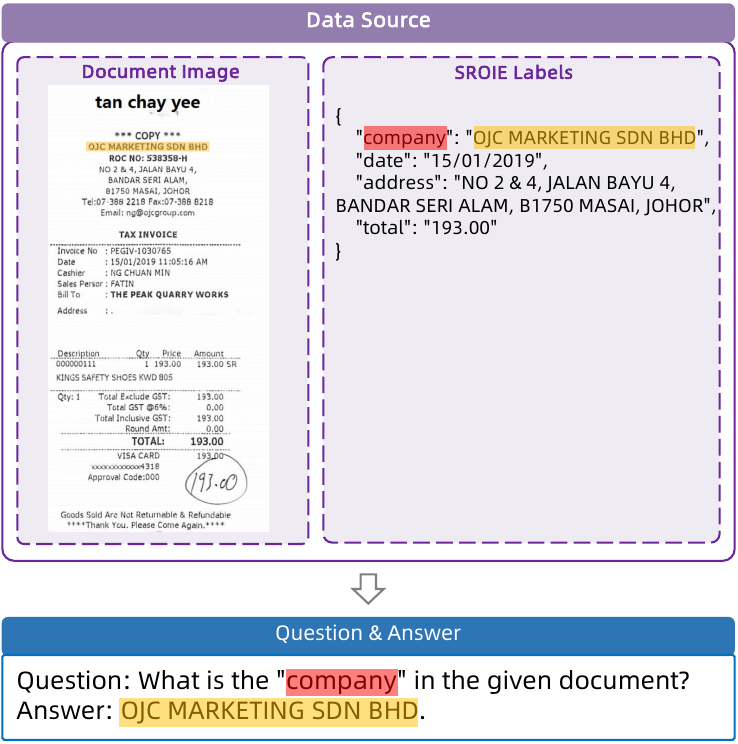}
    \caption{Evaluation data construction example of the QA for VIE task through entity annotations in VIE (SROIE\&CORD).}
    \label{fig:app_vie_sroie}
    \vspace{-4mm}
\end{figure*}

Different from the FUNSD dataset, the VIE annotations in the \textbf{CORD}~\cite{park2019cord} and \textbf{SROIE}~\cite{Huang_2019} datasets can be directly transformed to \{``entity type'':``entity text''\} as shown in \cref{fig:app_vie_sroie}.
Therefore, for entity annotations in CORD and SROIE, the question\&answer is constructed by directly asking the entity content.
Similar to the filter rules conducted on the FUNSD dataset, cases where a specific entity type appears multiple times in the document are filtered out.

In order to avoid the influence of randomness in generation on the evaluation, sampling methods are not used for any of the LLMs/MLLMs during testing.
The beam search with a beam size of 5 is employed for generation across all models.

%

\end{document}